\documentclass[runningheads]{llncs}

\usepackage{eccv}

\usepackage{eccvabbrv}

\usepackage{graphicx}
\usepackage{booktabs}
\usepackage[dvipsnames]{xcolor}

\usepackage{booktabs}
\usepackage{multicol}
\usepackage{multirow}
\usepackage{colortbl}
\usepackage{xcolor}
\usepackage{caption}
\usepackage{tabularray}
\usepackage{rotating}
\usepackage{makecell}
\usepackage{tabularx}
\usepackage{float}
\usepackage{adjustbox}
\usepackage{anyfontsize}
\usepackage{mathtools}
\usepackage{amsmath}
\usepackage{indentfirst}
\usepackage{float}

\usepackage{amssymb}
\usepackage{pifont}
\newcommand{\cmark}{\ding{51}}%
\usepackage{microtype}

\usepackage[accsupp]{axessibility}  

\usepackage{hyperref}

\usepackage{orcidlink}

\begin{document}

\hbadness=2000000000
\vbadness=2000000000
\hfuzz=100pt


\title{Diffusion Model is a Good Pose Estimator from 3D RF-Vision}

\titlerunning{mmDiff for 3D Pose Estimation}

\author{
Junqiao Fan\inst{1}\orcidlink{0000-0002-8465-5447}\index{Fan, Junqiao} \and
Jianfei Yang\inst{1,2}\thanks{J. Yang is the project lead.}\orcidlink{0000-0002-8075-0439}\index{Yang, Jianfei} \and
Yuecong Xu\inst{1}\orcidlink{0000-0002-4292-7379}\index{Xu, Yuecong} \and
Lihua Xie\inst{1}\orcidlink{0000-0002-7137-4136}\index{Xie, Lihua}
}

\authorrunning{J.~Fan et al.}

\institute{
School of Electrical and Electronic Engineering\\
\and
School of Mechanical and Aerospace Engineering\\
Nanyang Technological University, Singapore\\
\email{\{fanj0019, jianfei.yang, xuyu0014, elhxie\}@ntu.edu.sg}
\\
}

\maketitle
\begin{abstract}
Human pose estimation (HPE) from Radio Frequency vision (RF-vision) performs human sensing using RF signals that penetrate obstacles without revealing privacy (e.g., facial information). Recently, mmWave radar has emerged as a promising RF-vision sensor, providing radar point clouds by processing RF signals. However, the mmWave radar has a limited resolution with severe noise, leading to inaccurate and inconsistent human pose estimation. This work proposes mmDiff, a novel diffusion-based pose estimator tailored for noisy radar data. Our approach aims to provide reliable guidance as conditions to diffusion models. Two key challenges are addressed by mmDiff: (1) miss-detection of parts of human bodies, which is addressed by a module that isolates feature extraction from different body parts, and (2) signal inconsistency due to environmental interference, which is tackled by incorporating prior knowledge of body structure and motion. Several modules are designed to achieve these goals, whose features work as the conditions for the subsequent diffusion model, eliminating the miss-detection and instability of HPE based on RF-vision. Extensive experiments demonstrate that mmDiff outperforms existing methods significantly, achieving state-of-the-art performances on public datasets. \footnote[1]{The project page of mmDiff is \href{https://fanjunqiao.github.io/mmDiff-site/}{https://fanjunqiao.github.io/mmDiff-site/}.}
\end{abstract}    
\section{Introduction}
\label{sec:introduction}

Human pose estimation (HPE) has been a widely-studied computer vision task for predicting coordinates of human keypoints and generating human skeletons~\cite{andriluka20142d, martinez2017simple, zhou2023adapose}. It is a fundamental task for human-centered applications, such as augmented/virtual reality~\cite{mehta2017vnect, zhou2023metafi++, yang2022metafi}, rehabilitation~\cite{tao2020trajectory, ar2014computerized}, and human-robot interaction~\cite{gao2019dual,deng2023long}. Current HPE solutions mainly rely on RGB(D) cameras~\cite{chen2018cascaded, kocabas2020vibe}. Though demonstrating promising accuracy, camera-based pose estimators have intrinsic limitations under adverse environments, e.g., smoke, low illumination, and occlusion~\cite{wang2021human}. The privacy issue of cameras also hinders the viability of HPE in medical scenarios, e.g., rehabilitation systems in hospitals~\cite{an2022mri}.

Overcoming limitations of camera-based HPE, Radio-Frequency vision (RF-vision) has attracted surging attention for human sensing. The emerging mmWave radar technology presents a promising and feasible solution for HPE due to its price, portability, and energy efficiency~\cite{waldschmidt2021automotive}. Operated at the frequencies of 30-300 GHz~\cite{zhang2023survey}, commercial mmWave radars transmit and receive RF signals that penetrate human targets and occlusions. Radar point clouds (PCs) are further extracted as the salient target detection via monitoring signals' characteristic changes~\cite{rao2017introduction, xue2021mmmesh}. Therefore, the extracted PCs are more robust to adverse environments~\cite{zhao2018through} and reveal little privacy~\cite{yang2023mm}, inspiring accurate and privacy-preserving HPE~\cite{li2020capturing, yang2023mm, an2022fast, xue2021mmmesh}. However, due to the bandwidth and hardware limitations~\cite{prabhakara2023high}, radar PCs are sparse with limited geometric information~\cite {an2022fast}, leading to huge difficulties in achieving HPE. The sparse PCs are noisy in two aspects: (1) mmWave radar has a lower spatial resolution, leading to PC dispersion throughout the target area accompanied by ghost points caused by multi-path effect~\cite{sun20213drimr, prabhakara2023high}; (2) signal's specular reflection and interference~\cite{chen2022mmbody, bansal2020pointillism} further cause inconsistent sensing data, leading to occasional miss-detection of human parts.

\begin{figure}[!t]
\centering
\includegraphics[width=1.\linewidth]{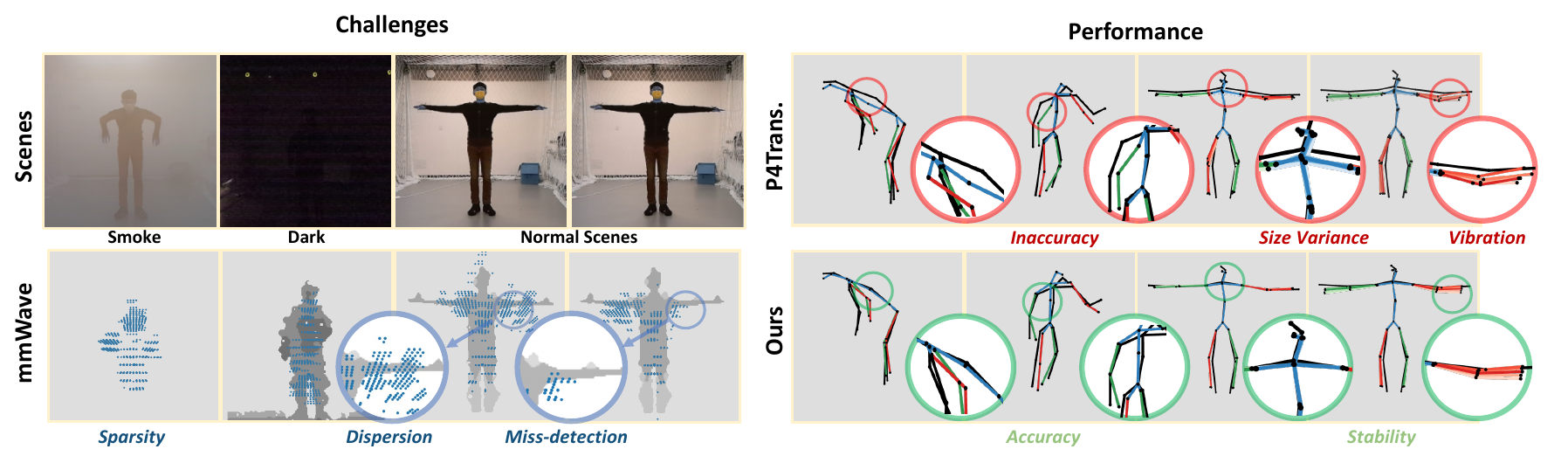}
\caption{Left: challenges of mmWave PCs. Right: the performance of existing SOTA (P4Transformer~\cite{fan2021point}) compared to ours. The GTs are black and predictions are colored. PC's sparsity and dispersion cause inaccurate spline and shoulder. Inconsistent PCs with occasional miss-detection further cause size variance and pose vibration. mmDiff proposes diffusion-based pose estimation with enhanced accuracy and stability.}

\label{Effect}
\end{figure}

To deal with sparse and noisy mmWave PCs, existing solutions mainly rely on kinds of data augmentations, e.g., multi-frame aggregation to enhance PC resolution~\cite{an2022fast, xue2021mmmesh}. For the feature extractor, they directly borrow Long Short-Term Memory (LSTM)~\cite{xue2021mmmesh} or transformer-based architectures~\cite{chen2022mmbody, yang2023mm} from existing RGB(D) HPE methods. However, these feature encoders are tailored for visual and language modalities, which struggle to handle noisy and inconsistent radar PCs~\cite{qi2020analyzing}. As shown in Figure~\ref{Effect} (right), the existing SOTA solution~\cite{fan2021point} still suffers from pose vibration and severe drift, achieving undesirable performance.

Denoising Diffusion Probabilistic Models (DDPMs)~\cite{ho2020denoising, song2020denoising}, also known as diffusion models, have demonstrated superior performance in various generative tasks, such as image generation and image restoration~\cite{hu2022global, zhu2023conditional, gong2023diffpose}. Diffusion models perform progressive noise elimination, transferring noisy distribution into desired target distribution~\cite{song2019generative}. Inspired by such capability, we aim to mitigate the noise of mmWave HPE, which motivates mmDiff, a diffusion-based pose estimator tailored for noisy radar PCs. Different from existing diffusion-based HPE using RGB(D), HPE using mmWave PCs confronts two key challenges: (1) extracting robust features from noisy PCs where miss-detection of human bodies may happen, and (2) overcoming signal inconsistency for stable HPE. For the first challenge, we propose to isolate the feature extraction for different body joints, so that occasional miss-detection would not affect the feature extraction of detected joints. Extracting features directly from local PCs also improves the feature resolution. For the inconsistency issue, prior knowledge of human body structure and motion can reduce unreasonable cases, achieving consistent feature learning. As the human structure has a size constraint where limb-length should remain constant~\cite{chen2021anatomy}, the limb-length can be additionally estimated to prevent pose variance. Moreover, inspired by human motion continuity~\cite{ramakrishna2013tracking} that discourages abrupt human behavior changes, historical poses can be leveraged for pose generation refinement, minimizing pose vibration.

To this end, mmDiff first designs a conditional diffusion model capable of injecting radar information as the guiding conditions. Four modules are designed to extract clean and consistent information from radar point clouds: (1) Global radar context is proposed to isolate the globally extracted features for different human joints using a transformer~\cite{vaswani2017attention}, which generates more robust joint-wise features to handle miss-detection. (2) Local radar context is proposed to extract local features around body joints with a local transformer, which performs point-level attention for higher resolution. (3) Structural human limb-length consistency is proposed to extract human limb-length as consistent patterns, which reduce limb-length variance. (4) Temporal motion consistency is proposed to learn smooth motion patterns from historical estimated poses, which avoid pose vibration. Experiments have shown a significant improvement in pose estimation accuracy using mmDiff compared to the state-of-the-art models. Meanwhile, the generated poses demonstrate comparable structural and motion stability, validating the effectiveness of our designed conditional modules.

In summary, our contributions are three-fold. First, we propose a novel diffusion-based HPE framework with sparse and noisy mmWave radar PCs. To the best of our knowledge, mmDiff is the first diffusion-based paradigm for mmWave radar-based HPE. Second, four modules are proposed to extract robust representations from the noisy and inconsistent radar PCs considering global radar context, local radar context, structural limb-length consistency, and temporal motion consistency, used as the conditions to guide the diffusion process. Finally, extensive experiments show our approach achieves state-of-the-art performance on two public datasets: mmBody~\cite{chen2022mmbody} and mm-Fi~\cite{yang2023mm}. 
\section{Related Work}
\label{sec:related_work}

\subsubsection{mmWave Human Pose Estimation.}
For decades, extensive works~\cite{andriluka20142d, martinez2017simple} have been focused on human pose estimation from RGB(D) images. Though achieving desirable accuracy, the major challenge faced by RGB(D) HPE is the performance drop under adverse environments and with self-occlusion~\cite{chen20232d}. Recently, commercial mmWave radar has been proven to extract sufficient information for human body reconstruction~\cite{xue2021mmmesh}, bringing the potential for mmWave HPE. Despite methods~\cite{lee2023hupr, xue2023towards} utilizing raw radar signals for mmWave HPE, point clouds-based methods~\cite{an2022fast,an2022mri,xue2021mmmesh,yang2023mm,chen2022mmbody} become popular for its format uniformity. Particularly, Xue et al.~\cite{xue2021mmmesh} utilizes the LSTM model with anchor-based local encoding to deal with the noisy nature of radar. An et al.~\cite{an2022fast} integrate point clouds of consecutive frames to handle the point cloud sparsity. Chen et al.~\cite{chen2022mmbody} and Yang et al.~\cite{yang2023mm} propose transformer-based benchmarks based on their dataset. However, the sparse and noisy radar point clouds still hinder the accuracy of HPE using existing non-parametric regression models.

\subsubsection{Diffusion Model for Human Pose Estimation.}
Diffusion models have been widely applied in image generation, such as image restoration~\cite{lugmayr2022repaint}, super-resolution~\cite{kawar2022denoising}, and text-to-image synthesis~\cite{rombach2022high}. Since the proposition of the denoising diffusion probabilistic model (DDPM)~\cite{ho2020denoising}, the diffusion model is extended to a wider range of generative applications, including pose/skeleton generation. DiffPose\cite{gong2023diffpose} formulate the 3D pose estimation problem as a pose generation task from low-determinant 2D poses to high-determinant 3D poses. Following DiffPose, Shan et al.~\cite{shan2023diffusion} proposes to generate multi-hypothesis poses for 3D pose ambiguity and Saadatnejad et al.~\cite{saadatnejad2023generic} proposes to predict the human poses in future time frames. Meanwhile, RGB-guided diffusion models for 2D and 3D human pose generation are also achieved~\cite{qiu2023learning, zhou2023diff3dhpe}. Nevertheless, existing methods focus on diffusion-based HPE from stable and informative modality sources, either well-estimated 2D poses or high-resolution RGB images. None of these works investigate how noisy and sparse mmWave radar point clouds are used to guide the pose generation.

\section{Methodology}
\label{sec:methodology}
\begin{figure*}[!ht]
\centering
\includegraphics[width=1.\linewidth]{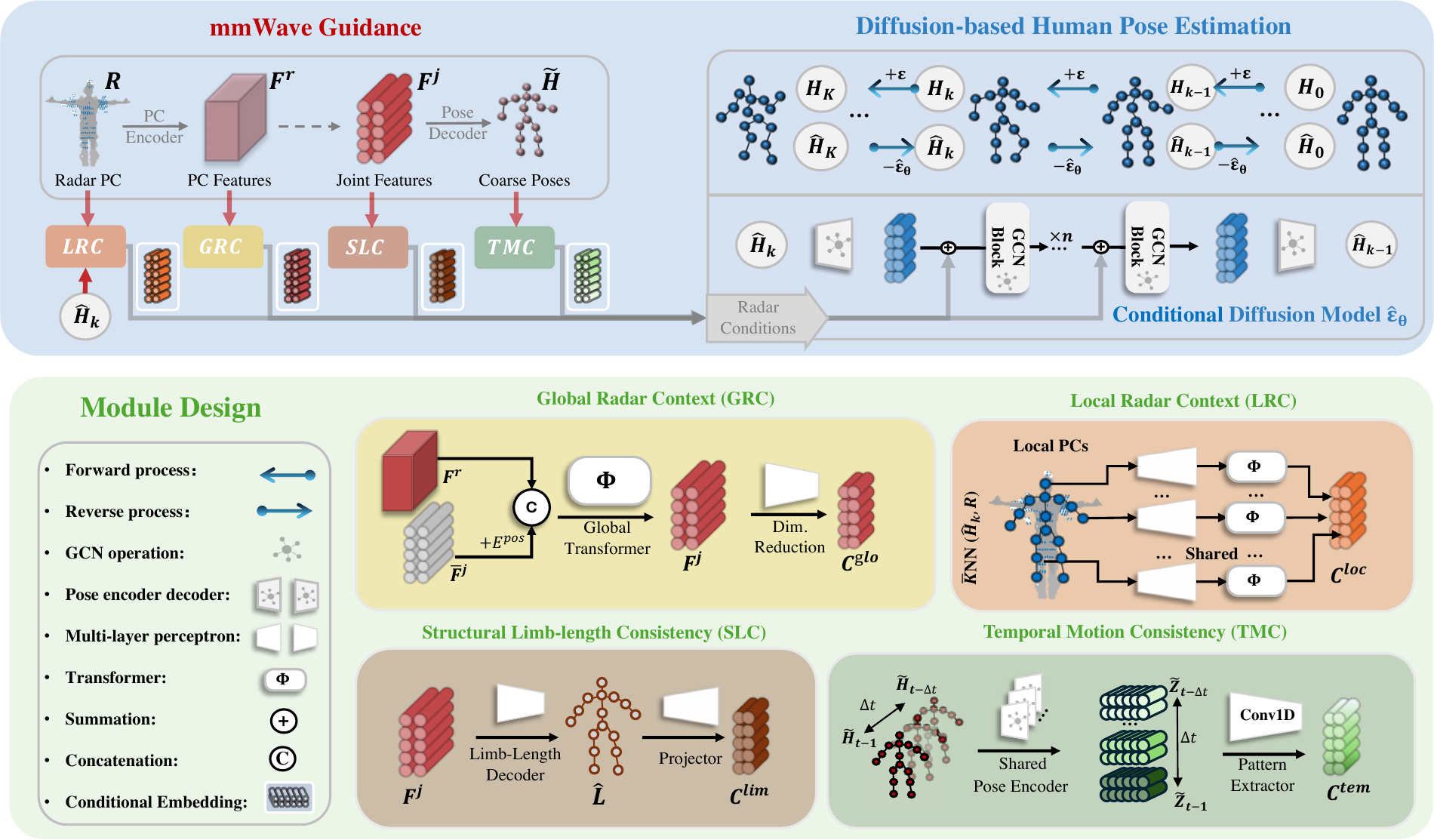}
\caption{mmDiff proposes a diffusion-based HPE model, using mmWave radar information as conditions. $k \in [0..K]$ denotes the diffusion step. Four modules are proposed as the more reliable guidance, addressing PCs' noise and inconsistency: GRC and LRC first extract robust global-local radar features, $C^{glo}$ and $C^{loc}$; SLC and TMC then extract consistent human structure and motion patterns, $C^{tem}$ and $C^{lim}$.}
\label{system architecture}
\end{figure*}

We address 3D HPE task with noisy and sparse PCs from mmWave radar. At each time frame $t \ge 0$, given the input 6-dimensional radar point cloud $R_t \in \mathbb{R}^{N \times 6}$, where $N$ denotes the number of detected points, our task is to estimate the ground-truth 17-joint human pose $H_t = \{h_t^1, h_t^2, ..., h_t^{17}\} \in \mathbb{R}^{17 \times 3}$. Apart from the Cartesian coordinate $\{x, y, z\}$, each detected radar point contains three attributes $\{v, E, A\}$, representing velocity, energy, and amplitude respectively.

To extract clean radar features and overcome mmWave signal inconsistency, we propose mmDiff as a diffusion-based paradigm for mmWave radar-based HPE. The overview of the architecture of mmDiff is presented in Figure~\ref{system architecture}. We start with an overall description of how human poses are generated by diffusion models using radar information as the guiding conditions in Section~\ref{diffusion HPE}, followed by a detailed illustration of the four modules for extracting robust representations from the noisy and inconsistent radar PCs in Section~\ref{global-local} and Section~\ref{spatial-temporal}. We omit the subscript $t$ denoting the time frame in subsequent discussions for clarity.

\subsection{Diffusion-based Human Pose Estimation}
\label{diffusion HPE}

Diffusion models~\cite{ho2020denoising, song2019generative} are a category of probabilistic generative models popular in various tasks, e.g., image generation. Given a noisy image from a noisy distribution, diffusion models can generate realistic image samples that match the natural image distribution, through iterative noise removal~\cite {lugmayr2022repaint, kawar2022denoising}. Extended to HPE, diffusion models can estimate the distribution of reasonable human poses for realistic pose generation. Particularly with noisy radar modality, miss-detected body joints can be accurately estimated by inferring from the detected ones, and inaccurate joints causing twisted human structures are potentially refined.

The diffusion-based HPE consists of two processes: (1) a forward process gradually generates noisier samples as the training guidance and (2) a reverse process learns to invert the forward diffusion. Modelled by a Markov chain of length $K$, the forward process starts from the ground truth pose $H_0$, iteratively samples a noisier pose $H_k$ by adding Gaussian noise $\epsilon \in \mathcal{N}(0, I)$:
\begin{equation}
    q\left(H_k \mid H_{k-1}\right) = \mathcal{N}\left(H_k \mid \sqrt{1-\beta_k} H_{k-1}, \beta_k I\right),
\label{eq：diff}
\end{equation}
where $k \in [0..K]$ refers to the diffusion step and $\beta_k$ refers to the noise scale. On the contrary, the reverse process starts from a noisy pose initialization $\hat{H}_K$ and progressively removes noises until $\hat{H}_0$ is generated. For each step $k$, a diffusion model $\hat{\varepsilon}_\theta$ is trained to identify the pose noise $\hat{\varepsilon}_k$ within $\hat{H}_k$, and remove it for more reliable pose $\hat{H}_{k-1}$:
\begin{align}
    \hat{\varepsilon}_k &= \hat{\varepsilon}_\theta(\hat{H}_k, k),\\
    \hat{H}_{k-1} &= (1-\beta_k)^{-1}(\hat{H}_{k} - \beta_k\hat{\varepsilon}_k).
\label{eq:diff23}
\end{align}

To better leverage the context information provided by radar's sensing data, we propose a set of conditions $C$ containing latent feature embeddings extracted from the mmWave modality, to guide each step of the reverse process: 
\begin{equation}
    \hat{\varepsilon}_k = \hat{\varepsilon}_\theta(\hat{H}_k, C, k).
\label{eq:diff4}
\end{equation}
As an implementation, we propose a conditional diffusion model, which injects the radar conditions $C$ in the latent feature space, using a Graph Convolution Network (GCN)~\cite{zhao2022graformer} as the backbone. The GCN takes the 17-joint human pose $H_k \in \mathbb{R}^{17 \times 3}$ as input, which is subsequently encoded into the latent pose embedding $Z_k \in \mathbb{R}^{17 \times 96}$ by a GCN encoder, fed into $n$ GCN blocks, and decoded into $\hat{\varepsilon}_k \in \mathbb{R}^{17 \times 3}$ by a GCN decoder. To inject radar conditions, we propose to add the conditional embeddings from $C$ to the pose feature $Z_k$ before each GCN block, which serves as extra information for more accurate noise estimation of $\hat{\varepsilon}_k$. To align features, the conditional embeddings are also projected to the $\mathbb{R}^{17 \times 96}$ feature space. In the following sections, we discuss in detail how to extract clean and consistent radar features that construct $C$. 

\subsection{Global-local Radar Context}
\label{global-local}
Accurate guidance for the diffusion model depends on robust radar feature extraction, which should carefully handle the miss-detection of human bodies. In this section, we first revisit the existing mmWave feature extraction paradigm. Then, we further propose to improve it with two modules: (1) Global Radar Context (GRC) extracting features from overall PCs that can handle miss-detection; and (2) Local Radar Context (LRC) extracting local PC features near body joints for higher resolution. 

\subsubsection{Revisit mmWave Feature Extraction.} Existing mmWave HPE paradigm applies encoder-decoder architectures to encode the radar PCs, $R \in \mathbb{R}^{N \times 6}$, into latent feature representations and decode them into human poses. Anchor-based methods~\cite{fan2021point, zhao2021point} are common options for PC encoders, where point anchors are first sampled from the Farthest Point Sampling algorithm and nearby PCs are extracted as the anchor features. As a result, the holistic radar PCs are encoded into PC features $F^{r} \in \mathbb{R}^{P \times 1024}$, where $P$ indicates the anchor number. To decode the PC features, a Multi-Layer Perceptron (MLP) is commonly applied as a dimension-reducing projection to generate the joint feature $F^{j} \in \mathbb{R}^{17 \times 1024}$, which is further decoded into human poses $\tilde{H} \in \mathbb{R}^{17 \times 3}$ by another MLP. However, occasional radar miss-detection causes uncertainty within the PC features $F^{r}$, while MLP-based projection can hardly identify miss-detected joints. Additionally, radar low-resolution PCs further impose noises. As a result, the estimated human joints are generally coarse. 

\subsubsection{Global Radar Context (GRC).} To handle occasional miss-detection, GRC is proposed to isolate feature extraction for different body joints, using global information from $F^{r}$ to construct more robust joint features $F^{j}$. From joint features, the diffusion model potentially identifies miss-detected joints and utilizes human prior knowledge for more accurate estimation. We exploit a Global-Transformer $\Phi^g$ to facilitate joint-wise feature extraction from the $F^{r}$. Following the $cls$ token design in ViT~\cite{dosovitskiy2020image}, we first randomly initialize a trainable joint feature template (with no information) $\Bar{F}^{j} \in \mathbb{R}^{17 \times 1024}$ and add it with positional embedding $E^{pos} \in \mathbb{R}^{17 \times 1024}$. Then, the joint feature template is concatenated with the PC feature $[\Bar{F}^{j}, F^{r}] \in \mathbb{R}^{(17+P) \times 1024}$ and fed into $\Phi^g$:
\begin{align}
F^{j}, F'^{r} = \Phi^{g}(\Bar{F}^{j}, F^{r}).
\label{eq:global1}
\end{align}
The output $F^{j} \in \mathbb{R}^{17 \times 1024}$ is selected as the joint feature and the rest $F'^{r}$ is ignored. Within the transformer, deep correlation is captured, not only within $F^{r}$ but also between $F^{j}$ and $F^{r}$. Each body joint performs individual feature extraction based on their correlation with the PC feature, so that features of detected joints are less affected by other undetected parts. Finally, as the 1024-dim $F^{j}$ feature space is too sparse for the diffusion latent condition, an MLP-based dimension-reduction function $g^{g}$ further condenses the extracted information within a $\mathbb{R}^{17 \times 64}$ conditional embedding:
\begin{equation}
C^{glo} = g^{g}(F^{j}).
\label{eq:global2}
\end{equation} 

\subsubsection{Local Radar Context (LRC).} LRC further performs local point-to-point self-attention to extract higher-resolution features from local PCs near body joints. To select local PCs, existing methods~\cite{xue2021mmmesh} utilize static joint anchors from coarsely-estimated human poses $\Tilde{H}$. However, the local joint features should dynamically reflect the joints' errors at different diffusion steps. Therefore, we propose dynamic joint anchors from intermediate diffusion poses $\hat{H}_{k}$ for PC selection. With $i \in [1,..,17]$ and each joint $\hat{h}_k^i$ from the $\hat{H}_{k}$ as an anchor, the $\bar{K}$-nearest-neighbors ($\bar{K}$NN) algorithm is applied to select $\bar{K}$ nearest points as local PCs, $\Bar{R}^{i}_{k} \in \mathbb{R}^{\bar{K} \times 6}$. Each $\Bar{R}^{i}_{k}$ is first encoded by a shared MLP $g^{l}$ into a $\mathbb{R}^{\bar{K} \times 64}$ embedding, and then fed into a shared small-scale local transformer $\Phi^{l}$ for point-to-point self-attention. Finally, average pooling is performed to generate $\mathbb{R}^{64}$ embeddings, which are further aggregated (concatenated) for every joint anchor $\tilde{h}^i$ as the conditional embedding:
\begin{equation}
    C^{loc} = \bigcup_{i \in [1,..,17]}\Phi^{l} \circ g^{l}(\Bar{R}^{i}_{k}).
\label{local}
\end{equation}

\subsection{Structural-motion Consistent Patterns}
\label{spatial-temporal}
Inconsistent radar signals such as occasional miss-detection lead to discontinuous and unstable pose estimation, such as variant limb-length, pose vibration, or inconsistent error frames. To mitigate such inconsistency, we further extract consistent human patterns based on human structure and motion prior knowledge: (1) Structural Limb-length Consistency (SLC) that extracts limb-length patterns to reduce limb-length variance, and (2) Temporal Motion Consistency (TMC) learns smooth motion patterns from historically estimated human poses.

\subsubsection{Structural Limb-Length Consistency (SLC).} SLC learns to extract a human-size indicator, the 16 limb-length $\hat{L}= \{\hat{l}^1, ..., \hat{l}^{16}\} \in \mathbb{R}^{16}$, where each limb-length measures the bone length connecting adjacent body joints. The extracted limb-length serves as a structural constraint to reduce the limb-length variance during the pose generation. Similar to decoding a pose, an MLP-based limb decoder $g_1^{lim}$ is first applied to decode the previously extracted global joint feature $F^{j}$ into predicted limb-length $\hat{L}$. Though with less dimension, the extracted limb-length contains physical meanings and thus is more consistent and stable. To ensure accurate limb-length decoding, a limb loss is designed to guide the training (details in Section~\ref{overall training}). To further project the estimated limb-length $\hat{L}$ into latent feature space, another MLP-based projector $g_2^{lim}$ then transforms the $\hat{L}$ into the $\mathbb{R}^{96}$ embedding, which is further broadcasted to different joints as the $\mathbb{R}^{17 \times 96}$ conditional embedding: 
\begin{equation}
    C^{lim} = g_2^{lim}(\hat{L}) = g_2^{lim} \circ g_1^{lim}(F^{j}).
\label{spatial}
\end{equation}

\subsubsection{Temporal Motion Consistency (TMC).} Inspired by the fact that human motion is generally stable and consistent~\cite{ramakrishna2013tracking}, multi-frame historical human poses can be utilized to extract the motion patterns in estimating the current pose. Such motion patterns provide temporal constraints for the diffusion model, which avoids error frames and pose vibration. As shown in Figure \ref{system architecture}, TMC extracts the latent motion patterns from a sequence of historical-estimated coarse poses $\{\tilde{H}_{t-\Delta t}, ..., \tilde{H}_{t-1}\}$, where $\Delta t$ is the number of historical frames. Firstly, a shared GCN encoder $g_1^{tem}$ is applied to convert the pose sequence into feature embedding sequence $\{\Tilde{Z}_{(t-\Delta t)}, ..., \Tilde{Z}_{(t-1)}\} \in \mathbb{R}^{\Delta t \times (17 \times 96)}$. Then, a 1D-convolution-based pattern extractor $g_2^{tem}$ is applied to extract the motion information along the temporal dimension, which generates a $\mathbb{R}^{17 \times 96}$ temporal embedding as $C^{tem}$:
\begin{equation}
    C^{tem} = g_2^{tem}\left(\bigcup_{i \in [1..\Delta t]} \Tilde{Z}_{(t-i)}\right) = g_2^{tem}\left(\bigcup_{i \in [1..\Delta t]} g_1^{tem}(\Tilde{H}_{t-i})\right).
\label{temporal}
\end{equation}
The reason for using 1D convolution is two-fold: (1) smooth motion features can be extracted by potentially averaging pose features of historical frames, which avoids pose vibration; and (2) the motion trend of the on-performing actions is potentially extracted to avoid inconsistent error frames. For example, increasing $z$ values of the hand’s location are expected when performing the `raising hand'.

\subsection{Overall Learning Objective}
\label{overall training}
As feature extraction from global radar PCs is computation-consuming, we divide the training process into two phases. Phase one facilitates the extraction of the global joint features $F^j$ and coarse estimation of human poses $\tilde{H}$. The GRC, PC encoder (from an off-the-shelf mmWave HPE network), and an MLP-based pose decoder are trained together. The learning objective of the phase one is minimizing the $\mathcal{L}_2$ pose regression loss $\mathcal{L}_{\text{joint}}$:
\begin{equation}
    \mathcal{L}_{\text{joint}}={E}_{i \sim[1,17]}|| h^i-\tilde{h}^i \|_2^2.
\label{loss1}
\end{equation} 
In phase two, the remaining three conditional modules and the diffusion model are trained together, with the phase one parameters frozen. The extracted $F^j$ and $\tilde{H}$ serve as the input of structural-motion consistency modules, and $\tilde{H}$ initialize $\hat{H}_K$ for the reverse diffusion process. The learning objective is minimizing $\mathcal{L}_{\text {diff}}$, which is the diffusion learning objective following DDPM~\cite{ho2020denoising}. To ensure accurate limb-length estimation, an $\mathcal{L}_1$ limb regression loss is further designed and integrated:
\begin{align}
\begin{split}
    \mathcal{L}_{\text {diff}}=\mathbb{E}_{k \sim[1, T]} \mathbb{E}_{\varepsilon_k \sim \mathcal{N}(0, I)}|| \varepsilon_k - \hat{\varepsilon}_\theta(H_k, k, C)\|^2_2 \\ + \lambda * \mathbb{E}_{i \sim[1, 16]}|l^i-\hat{l}^i|_1,
\end{split}
\label{loss2}
\end{align}
where $\lambda$ is a weighting parameter and each $l^i$ is the ground truth limb-length calculated from ground truth pose $H$.

\section{Experiments}
\label{sec:experiments}


\subsection{Experiment Setup}
\label{Experiment Setup}

\subsubsection{Datasets.}
mmBody~\cite{yang2023mm} studies the robustness of human sensing with various sensors: RGB, Depth, and mmWave radar. Human skeletons are annotated by the MoCap system.
Models are set to train on data collected from 2 standard scenes (Lab1 and Lab2), and tested on 3 basic scenes and 4 adverse scenes (including unseen subjects).
Meanwhile, mm-Fi~\cite{yang2023mm} is a larger-scale HPE dataset using a lower-bandwidth mmWave radar with sparser PCs. The skeleton annotations are rather unstable compared to the MoCap annotations, as obtained by RGB using the pretrained HRNet-w48~\cite{sun2019deep}. Three data-splitting methods are used with a train-test split ratio of 4:1: random, cross-subject, and cross-environment.


\subsubsection{Implementation Details.}
Our methods are trained for 100 epochs with a batch size of 1024. The Adam algorithm~\cite{kingma2014adam} is used for optimization, with the learning rate as $2e-5$, the gradient clip as 1.0, and the momentum as 0.9. The forward/reverse diffusion process is set as $K=25$ steps with a constant $\beta$ sampling of 0.001. An average of 5 hypotheses is recorded for a fair comparison with non-diffusion methods. We choose Point4D~\cite{fan2021point} as the GRC's PC encoder for mmBody and PointTransformer~\cite{zhao2021point} for mm-Fi. We set $\Bar{K}=50$ for $\Bar{K}$NN algorithm for LRC, but due to insufficient radar points ($N<100$), the LRC module is neglected for mm-Fi. We further set $\Delta t = 8$ for TMC and $\lambda = 5$ for SLC. The hyper-parameters are obtained empirically, more precise hyper-parameter tuning tricks such as the Bayesian optimization could lead to better results. 


\begin{table*}[!t]
\footnotesize
\centering
\caption{Quantitative results on mmBody~\cite{chen2022mmbody}, evaluated by MPJPE in white and PA-MPJPE in grey. * indicates diffusion-based methods. G, L, T, and S denote GRC, LRC, TMC, and SLC respectively. Bold is the best.}
\setlength{\belowcaptionskip}{0.1cm}
\resizebox{.95\textwidth}{!}{
  \fontsize{12}{14}\selectfont
    \begin{tabular}{l|cccccc|cccccccc|cc}
    \toprule
    \multicolumn{1}{l|}{\multirow{2}[2]{*}{\textbf{Methods}}} & \multicolumn{6}{c|}{\textbf{Basic Scenes}}    & \multicolumn{8}{c|}{\textbf{Adverse Environment}}             & \multicolumn{2}{c}{} \\
          & \multicolumn{2}{c}{\textbf{Lab1}} & \multicolumn{2}{c}{\textbf{Lab2}} & \multicolumn{2}{c|}{\textbf{Furnished}} & \multicolumn{2}{c}{\textbf{Rain}} & \multicolumn{2}{c}{\textbf{Smoke}} & \multicolumn{2}{c}{\textbf{Dark}} & \multicolumn{2}{c|}{\textbf{Occlusion}} & \multicolumn{2}{c}{\textbf{Average}} \\
    \midrule[1.5pt]\midrule
    RGB~\cite{chen2022mmbody}   & 74    & \cellcolor[rgb]{ .851,  .851,  .851}/ & 73    & \cellcolor[rgb]{ .851,  .851,  .851}/ & 71    & \cellcolor[rgb]{ .851,  .851,  .851}/ & 80    & \cellcolor[rgb]{ .851,  .851,  .851}/ & 86    & \cellcolor[rgb]{ .851,  .851,  .851}/ & 105    & \cellcolor[rgb]{ .851,  .851,  .851}/ & /     & \cellcolor[rgb]{ .851,  .851,  .851}/ & 81    & \cellcolor[rgb]{ .851,  .851,  .851}/ \\
    Depth~\cite{chen2022mmbody} & 55    & \cellcolor[rgb]{ .851,  .851,  .851}/ & 39    & \cellcolor[rgb]{ .851,  .851,  .851}/ & 55    & \cellcolor[rgb]{ .851,  .851,  .851}/ & 86    & \cellcolor[rgb]{ .851,  .851,  .851}/ & 243   & \cellcolor[rgb]{ .851,  .851,  .851}/ & 51    & \cellcolor[rgb]{ .851,  .851,  .851}/ & /     & \cellcolor[rgb]{ .851,  .851,  .851}/ & 88    & \cellcolor[rgb]{ .851,  .851,  .851}/ \\
    RGB(D)~\cite{chen2022mmbody} & 58    & \cellcolor[rgb]{ .851,  .851,  .851}/ & 34    & \cellcolor[rgb]{ .851,  .851,  .851}/ & 54    & \cellcolor[rgb]{ .851,  .851,  .851}/ & 95    & \cellcolor[rgb]{ .851,  .851,  .851}/ & 154   & \cellcolor[rgb]{ .851,  .851,  .851}/ & 58   & \cellcolor[rgb]{ .851,  .851,  .851}/ & /     & \cellcolor[rgb]{ .851,  .851,  .851}/ & 75    & \cellcolor[rgb]{ .851,  .851,  .851}/ \\
    \midrule
    MM-Mesh~\cite{xue2021mmmesh} & 95.1  & \cellcolor[rgb]{ .851,  .851,  .851}69.48 & 87.87 & \cellcolor[rgb]{ .851,  .851,  .851}77.3 & 93.28 & \cellcolor[rgb]{ .851,  .851,  .851}73.14 & 106.9 & \cellcolor[rgb]{ .851,  .851,  .851}72.38 & 106.7 & \cellcolor[rgb]{ .851,  .851,  .851}76.52 & 83.54 & \cellcolor[rgb]{ .851,  .851,  .851}64.19 & 85.55 & \cellcolor[rgb]{ .851,  .851,  .851}62.5 & 94.13 & \cellcolor[rgb]{ .851,  .851,  .851}70.79 \\
    P4transformer~\cite{fan2021point} & 69.35 & \cellcolor[rgb]{ .851,  .851,  .851}54.45 & 73.40 & \cellcolor[rgb]{ .851,  .851,  .851}66.39 & 75.28 & \cellcolor[rgb]{ .851,  .851,  .851}55.77 & 86.83 & \cellcolor[rgb]{ .851,  .851,  .851}65.71 & 89.82 & \cellcolor[rgb]{ .851,  .851,  .851}69.73 & 73.48 & \cellcolor[rgb]{ .851,  .851,  .851}54.52 & 78.56 & \cellcolor[rgb]{ .851,  .851,  .851}57.36 & 78.10 & \cellcolor[rgb]{ .851,  .851,  .851}60.56 \\
    PoseFormer~\cite{zheng20213d} & 64.53 & \cellcolor[rgb]{ .851,  .851,  .851}51.61 & 70.17 & \cellcolor[rgb]{ .851,  .851,  .851}63.26 & 69.71 & \cellcolor[rgb]{ .851,  .851,  .851}51.04 & 77.49 & \cellcolor[rgb]{ .851,  .851,  .851}59.03 & 84.82 & \cellcolor[rgb]{ .851,  .851,  .851}63.57 & 69.88 & \cellcolor[rgb]{ .851,  .851,  .851}50.46 & 73.52 & \cellcolor[rgb]{ .851,  .851,  .851}53.53 & 72.87 & \cellcolor[rgb]{ .851,  .851,  .851}56.07 \\
    DiffPose*~\cite{gong2023diffpose} & 66.43 & \cellcolor[rgb]{ .851,  .851,  .851}52.56 & 68.36 & \cellcolor[rgb]{ .851,  .851,  .851}65.69 & 69.78 & \cellcolor[rgb]{ .851,  .851,  .851}51.29 & 77.77 & \cellcolor[rgb]{ .851,  .851,  .851}62.60 & 89.01 & \cellcolor[rgb]{ .851,  .851,  .851}69.34 & 67.27 & \cellcolor[rgb]{ .851,  .851,  .851}49.90 & 74.52 & \cellcolor[rgb]{ .851,  .851,  .851}56.20 & 73.31 & \cellcolor[rgb]{ .851,  .851,  .851}58.23 \\
    \midrule
    mmDiff(G)* & 61.00 & \cellcolor[rgb]{ .851,  .851,  .851}49.19 & \textbf{67.79} & \cellcolor[rgb]{ .851,  .851,  .851}62.45 & 69.83 & \cellcolor[rgb]{ .851,  .851,  .851}51.47 & 77.39 & \cellcolor[rgb]{ .851,  .851,  .851}60.48 & 81.41 & \cellcolor[rgb]{ .851,  .851,  .851}64.44 & 68.83 & \cellcolor[rgb]{ .851,  .851,  .851}49.82 & 70.64 & \cellcolor[rgb]{ .851,  .851,  .851}52.79 & 70.99 & \cellcolor[rgb]{ .851,  .851,  .851}55.80 \\
    mmDiff(G,L)* & 61.11 & \cellcolor[rgb]{ .851,  .851,  .851}49.41 & 69.06 & \cellcolor[rgb]{ .851,  .851,  .851}63.14 & 68.17 & \cellcolor[rgb]{ .851,  .851,  .851}50.60 & 73.70 & \cellcolor[rgb]{ .851,  .851,  .851}58.45 & 82.26 & \cellcolor[rgb]{ .851,  .851,  .851}64.30 & 66.06 & \cellcolor[rgb]{ .851,  .851,  .851}48.56 & 68.18 & \cellcolor[rgb]{ .851,  .851,  .851}50.8 & 69.79 & \cellcolor[rgb]{ .851,  .851,  .851}55.04 \\
    mmDiff(G,L,T)* & 59.90 & \cellcolor[rgb]{ .851,  .851,  .851}\textbf{47.81} & 68.12 & \cellcolor[rgb]{ .851,  .851,  .851}62.02 & \textbf{66.98} & \cellcolor[rgb]{ .851,  .851,  .851}\textbf{48.63} & 74.84 & \cellcolor[rgb]{ .851,  .851,  .851}\textbf{58.13} & 80.95 & \cellcolor[rgb]{ .851,  .851,  .851}63.44 & 65.25 & \cellcolor[rgb]{ .851,  .851,  .851}\textbf{47.26} & 68.05 & \cellcolor[rgb]{ .851,  .851,  .851}50.4 & 69.16 & \cellcolor[rgb]{ .851,  .851,  .851}53.96 \\
    mmDiff(G,L,T,S)* & \textbf{59.52} & \cellcolor[rgb]{ .851,  .851,  .851}47.85 & 69.36 & \cellcolor[rgb]{ .851,  .851,  .851}\textbf{61.23} & 67.15 & \cellcolor[rgb]{ .851,  .851,  .851}49.19 & \textbf{71.03} & \cellcolor[rgb]{ .851,  .851,  .851}58.40 & \textbf{76.92} & \cellcolor[rgb]{ .851,  .851,  .851}\textbf{62.25} & \textbf{65.08} & \cellcolor[rgb]{ .851,  .851,  .851}47.47 & \textbf{67.47} & \cellcolor[rgb]{ .851,  .851,  .851}\textbf{49.54} & \textbf{68.08} & \cellcolor[rgb]{ .851,  .851,  .851}\textbf{53.71} \\
    \bottomrule
    \end{tabular}%
    }
    \label{tab:maintable}
    
\end{table*}%

\begin{table*}[t!]
\footnotesize
\centering
\caption{Quantitative results on mm-Fi~\cite{yang2023mm}.}
\setlength{\belowcaptionskip}{0.1cm}
\resizebox{.8\textwidth}{!}{%
    \begin{tabular}{p{10em}|cc|cc|cc}
    \toprule
    \multicolumn{1}{l|}{\multirow{2}[2]{*}{\textbf{Methods}}} & \multicolumn{2}{c|}{\textbf{Random}} & \multicolumn{2}{c|}{\textbf{Cross-Subject}} & \multicolumn{2}{c}{\textbf{Cross-Environment}} \\
    \multicolumn{1}{c|}{} & \multicolumn{1}{c}{\textbf{MPJPE}} & \multicolumn{1}{c|}{\textbf{PA-MPJPE}} & \multicolumn{1}{c}{\textbf{MPJPE}} & \multicolumn{1}{c|}{\textbf{PA-MPJPE}} & \multicolumn{1}{c}{\textbf{MPJPE}} & \multicolumn{1}{c}{\textbf{PA-MPJPE}} \\
    \midrule[1.5pt] \midrule
    PointTransformer~\cite{zhao2021point} & $73.09_{\pm 2.70}$ & $55.60_{\pm 1.40}$ & $75.96_{\pm 6.90}$ & $58.70_{\pm 4.30}$ & $88.28_{\pm 4.50}$ & $68.79_{\pm 2.79}$ \\
    Diffpose*~\cite{gong2023diffpose} & $73.44_{\pm 0.29}$ & $56.83_{\pm 0.25}$ & $70.31_{\pm 0.27}$ & $54.12_{\pm 0.31}$ & $86.35_{\pm 0.06}$ & $66.87_{\pm 0.17}$ \\
    mmDiff(G)* & $68.62_{\pm 0.06}$ & $53.11_{\pm 0.05}$ & $68.46_{\pm 0.06}$ & $52.55_{\pm 0.05}$  & $85.63_{\pm 0.53}$ & $66.43_{\pm 0.28}$ \\
    mmDiff(G,S)* & $65.72_{\pm 0.08}$ & $50.72_{\pm 0.01}$ & $67.18_{\pm 0.18}$ & $51.85_{\pm 0.05}$ & $83.39_{\pm 0.17}$  & $64.61_{\pm 0.40}$ \\
    mmDiff(G,S,T)* & $\textbf{65.26}_{\pm \textbf{0.11}}$ & $\textbf{50.35}_{\pm \textbf{0.09}}$ & $\textbf{65.62}_{\pm \textbf{0.24}}$ & $\textbf{50.23}_{\pm \textbf{0.24}}$ & $\textbf{82.73}_{\pm \textbf{0.62}}$ & $\textbf{63.87}_{\pm \textbf{0.26}}$ \\
    \bottomrule
    \end{tabular}%
    }    
    \label{tab:mmfi}
\end{table*}%

\subsubsection{Compared Methods.} 
(1) RGB, Depth, and RGB(D)~\cite{chen2022mmbody} are benchmarks using different modalities. (2) mmWave methods: P4Transformer~\cite{fan2021point} is the benchmark on mmBody containing Point4D Convolution as the PC encoder and a transformer for self-attention; PointTransformer~\cite{zhao2021point} is the benchmark on mm-Fi designed to handle the sparser PCs. mmMesh~\cite{xue2021mmmesh} is implemented as an extra mmWave HPE baseline on mmBody. (3) SOTA camera-based HPE methods: camera-based 3D HPE methods generally perform pose lifting from 2D poses to 3D poses. To modify them to perform radar-based HPE, we train the models to perform pose refinements, from coarse poses $\tilde{H}$ to clean 3D poses. PoseFormer~\cite{zheng20213d} is the transformer-based method with temporal-spatial attention. DiffPose~\cite{gong2023diffpose} is the diffusion-based method using SOTA graph-based GraFormer~\cite{zhao2022graformer} as backbones.

\subsubsection{Evaluation Metric.}
Two evaluation metrics are adopted following~\cite{ionescu2013human3}: (1) Mean Per Joint Position Error (MPJPE (mm)): the average joint error between ground truth and prediction (after pelvis alignment); and (2) Procrustes Analysis MPJPE (PA-MPJPE (mm)): procrustes methods (translation, rotation, and scaling) are performed before error calculation.

\begin{figure}[!t]
\centering
\includegraphics[width=.85\linewidth]{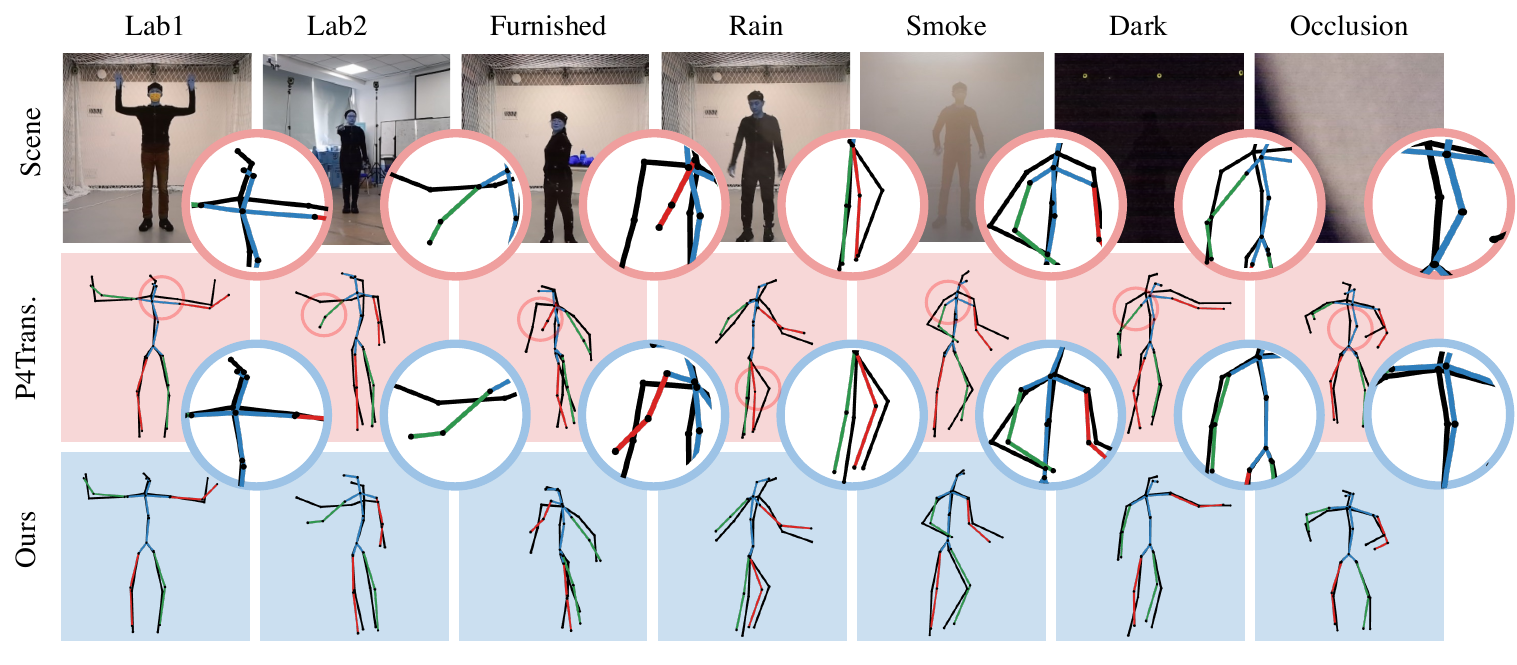}
\caption{Qualitative visualization of the estimated poses on mmBody dataset. mmDiff demonstrates higher keypoint accuracy. The GTs are black and predictions are colored.}
\label{exp1} 
\end{figure}

\begin{table}[!b]
\centering
\caption{Ablation studies of proposed modules on mmBody.}
\renewcommand{\arraystretch}{1.5}
\adjustbox{width=.85\linewidth}{

    \begin{tabular}{ll|p{3em}p{3em}|p{3em}p{3em}p{3em}|p{3em}p{3em}p{3em}|p{3em}p{3em}p{3em}p{3em}|p{3em}}
          &  & \multicolumn{2}{c|}{Diffusion Model} & \multicolumn{3}{c|}{Context Modules} & \multicolumn{3}{c|}{Consistency Modules} & \multicolumn{4}{c|}{Modules Elimination} & \multicolumn{1}{c}{Overall} \\
    \midrule[1.5pt] 
    \multirow{5}{*}{Modules} & Diffusion &       & \multicolumn{1}{c|}{\cmark} & \multicolumn{1}{c}{\cmark} & \multicolumn{1}{c}{\cmark} & \multicolumn{1}{c|}{\cmark} & \multicolumn{1}{c}{\cmark} & \multicolumn{1}{c}{\cmark} & \multicolumn{1}{c|}{\cmark} & \multicolumn{1}{c}{\cmark} & \multicolumn{1}{c}{\cmark} & \multicolumn{1}{c}{\cmark} & \multicolumn{1}{c|}{\cmark} & \multicolumn{1}{c}{\cmark} \\
          & Global &       &       & \multicolumn{1}{c}{\cmark} &       & \multicolumn{1}{c|}{\cmark} &       &       &       &       & \multicolumn{1}{c}{\cmark} & \multicolumn{1}{c}{\cmark} & \multicolumn{1}{c|}{\cmark} & \multicolumn{1}{c}{\cmark} \\
          & Local &       &       &       & \multicolumn{1}{c}{\cmark} & \multicolumn{1}{c|}{\cmark} &       &       &       & \multicolumn{1}{c}{\cmark} &       & \multicolumn{1}{c}{\cmark} & \multicolumn{1}{c|}{\cmark} & \multicolumn{1}{c}{\cmark} \\
          & Limb  &       &       &       &       &       & \multicolumn{1}{c}{\cmark} &       & \multicolumn{1}{c|}{\cmark} & \multicolumn{1}{c}{\cmark} & \multicolumn{1}{c}{\cmark} &       & \multicolumn{1}{c|}{\cmark} & \multicolumn{1}{c}{\cmark} \\
          & Temporal &       &       &       &       &       &       & \multicolumn{1}{c}{\cmark} & \multicolumn{1}{c|}{\cmark} & \multicolumn{1}{c}{\cmark} & \multicolumn{1}{c}{\cmark} & \multicolumn{1}{c}{\cmark} &       & \multicolumn{1}{c}{\cmark} \\
    \midrule[1.5pt]\midrule
    \multirow{2}{*}{Average} & MPJPE & \multicolumn{1}{c}{78.10} & \multicolumn{1}{c|}{73.31} & \multicolumn{1}{c}{70.99} & \multicolumn{1}{c}{70.43} & \multicolumn{1}{c|}{69.79} & \multicolumn{1}{c}{69.85} & \multicolumn{1}{c}{71.71} & \multicolumn{1}{c|}{69.46} & \multicolumn{1}{c}{68.83} & \multicolumn{1}{c}{68.81} & \multicolumn{1}{c}{69.45} & \multicolumn{1}{c|}{69.16} & \multicolumn{1}{c}{\textbf{68.08}} \\
          & PA-MPJPE & \multicolumn{1}{c}{60.56} & \multicolumn{1}{c|}{58.23} & \multicolumn{1}{c}{55.80} & \multicolumn{1}{c}{55.45} & \multicolumn{1}{c|}{55.04} & \multicolumn{1}{c}{54.90} & \multicolumn{1}{c}{57.06} & \multicolumn{1}{c|}{54.70} & \multicolumn{1}{c}{54.18} & \multicolumn{1}{c}{54.58} & \multicolumn{1}{c}{54.53} & \multicolumn{1}{c|}{53.96} & \multicolumn{1}{c}{\textbf{53.71}} \\
    \bottomrule
    \end{tabular}%
   }
\label{tab:ablation}
\end{table}

\subsection{Overall Result}
\label{Overall Result}
\subsubsection{Performance on mmBody.} As shown in Table \ref{tab:maintable}, mmWave-based methods have better robustness for cross-domain scenes and adverse environments. Particularly, our proposed mmDiff demonstrates superior results compared to related methods on the mmBody dataset. Compared to the SOTA mmWave HPE method, our method mmDiff(G,L,T,S) outperforms the P4Transformer~\cite{fan2021point} by 12.8\% (MPJPE) and 11.3\% (PA-MPJPE). For adverse environments, the improvement is more significant with 14.7\% (MPJPE) and 12.0\% (PA-MPJPE), because radar signals get noisier due to the specular reflection and mmDiff has improved noise-handling capability. Compared to Diffpose\cite{gong2023diffpose} and PoseFormer~\cite{zheng20213d} that utilize SOTA 3D HPE methods for pose refinement,  mmDiff(G,L,T,S) still outperforms by 6.6\% (MPJPE) and 4.2\% (PA-MPJPE), demonstrating the proposed modules are dedicated to handle noisy and sparse radar modalities. Furthermore, mmDiff enables better mmWave-based performance compared to RGB-based methods under all scenes and RGB(D)-based methods under adverse environments. Qualitatively, we observe more accurate human poses as illustrated in Figure \ref{exp1}.   

\subsubsection{Performance on mm-Fi.} As shown in Table~\ref{tab:mmfi}, the generalizability of our proposed mmDiff is further explored on mm-Fi dataset. We compare the proposed method with the benchmark PointTransformer~\cite{zhao2021point} and the SOTA DiffPose pose refinement~\cite{gong2023diffpose}. Compared with the benchmark, our proposed method reduces the pose estimation error by 6.29\% to 13.61\% (MPJPE) and 7.15\% to 14.42\% (PA-MPJPE). Though the vanilla DiffPose for pose refinement can improve the result, the performance is further improved by 4.19\% to 11.14\% (MPJPE) and 4.49\% to 11.40\% (PA-MPJPE) with the integration of the radar context module and the consistency modules, further demonstrating their effectiveness. Moreover, using cross-subject splitting, our method shows comparable results to random splitting. As our model explicitly learns the human structural and motion patterns from the radar signals, these patterns potentially generalize for unseen subjects. Our method also demonstrates cross-domain ability as the performance steadily improves with the integration of different modules. 

\begin{figure}[!t]
\centering
\includegraphics[width=.9\linewidth]{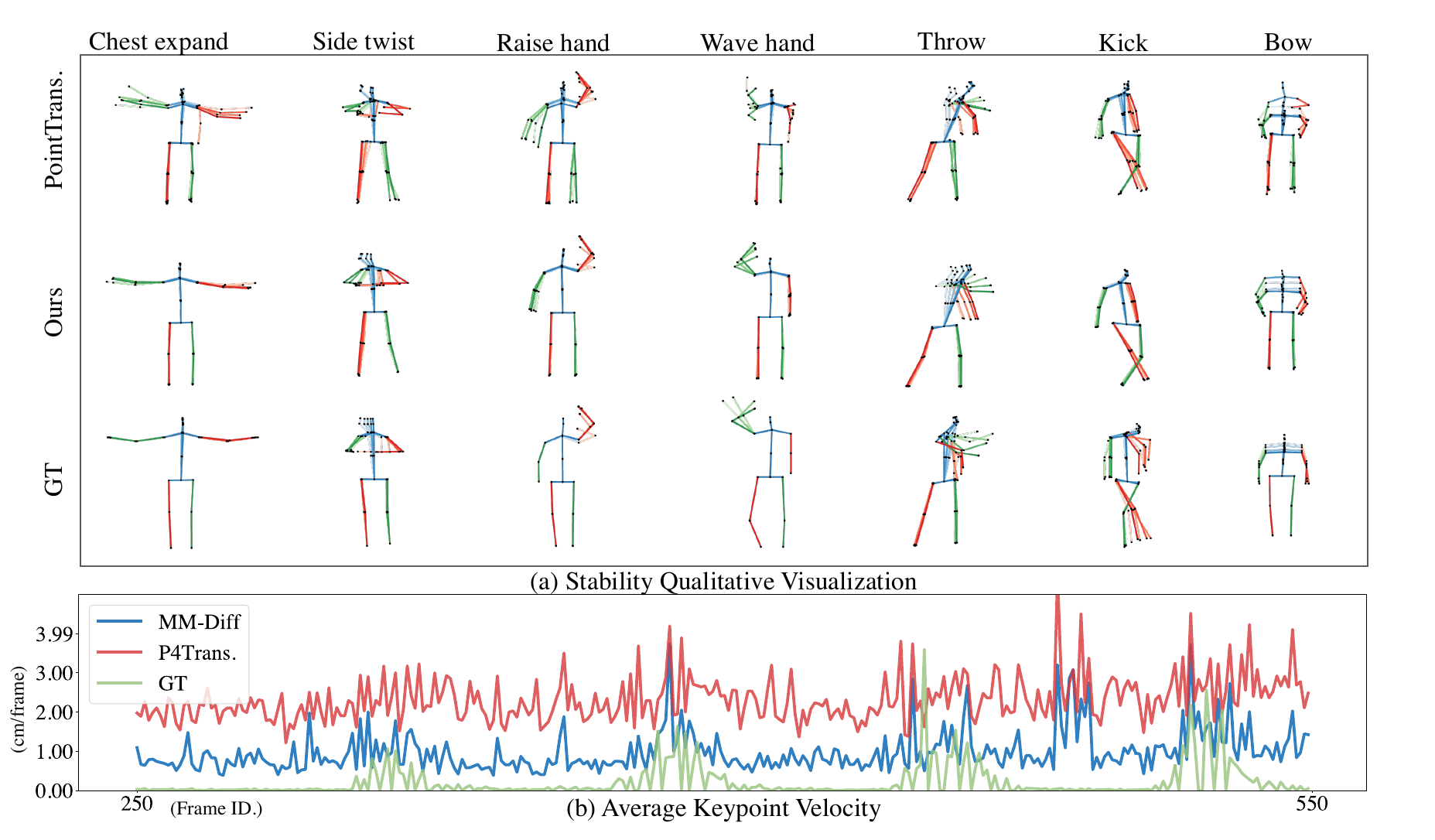}
\caption{(a) shows the pose motion stability on mm-Fi by plotting 5 consecutive frames of poses. mmDiff shows more consistent motion patterns (zoom in for details). (b) shows the motion energy levels on mmBody, where lower AKV indicates better stability.}
\label{exp2}
\end{figure}

\begin{table}[!b]
\centering
\caption{Ablation studies of the detailed designs in global-local radar context modules.}
\renewcommand{\arraystretch}{1.0}
\adjustbox{width=.75
\linewidth}{
    \begin{tabular}{ll|cccp{4em}ll|ccc}
    \cmidrule{1-5}\cmidrule{7-11}
          \multicolumn{5}{c}{Global Radar Context} &       & \multicolumn{5}{c}{Local Radar Context} \\
\cmidrule[1.5pt]{1-5}\cmidrule[1.5pt]{7-11}
\cmidrule{1-5}\cmidrule{7-11}    
\multicolumn{1}{l}{\multirow{3}{*}{Modules}} & Diffusion & \multicolumn{1}{c}{\cmark} & \multicolumn{1}{c}{\cmark} & \multicolumn{1}{c}{\cmark} &       & \multicolumn{1}{l}{\multirow{3}{*}{Modules}} & Diffusion & \multicolumn{1}{c}{\cmark} & \multicolumn{1}{c}{\cmark} & \multicolumn{1}{c}{\cmark} \\
          & PC Features $F^r$&       & \multicolumn{1}{c}{\cmark} &       &       &       & Static Anchors &       & \multicolumn{1}{c}{\cmark} &  \\
          & Joint Features $F^j$&       &       & \multicolumn{1}{c}{\cmark} &       &       & Dynamic Anchors &       &       & \multicolumn{1}{c}{\cmark} \\
\cmidrule{1-5}\cmidrule{7-11}    \multirow{2}{*}{Average} & MPJPE & 73.31 & 72.50 & \textbf{70.99} &       & \multirow{2}{*}{Average} & MPJPE & 73.31 & 71.05      & \textbf{70.43} \\
          & PA-MPJPE & 58.23 & 58.03 & \textbf{55.80} &       &       & PA-MPJPE & 58.23 &    56.06   & \textbf{55.45} \\
\cmidrule{1-5}\cmidrule{7-11}    
\end{tabular}%
}
\label{tab:global-local}
\end{table}

\begin{figure}[!t]
\centering
\includegraphics[width=.8\linewidth]{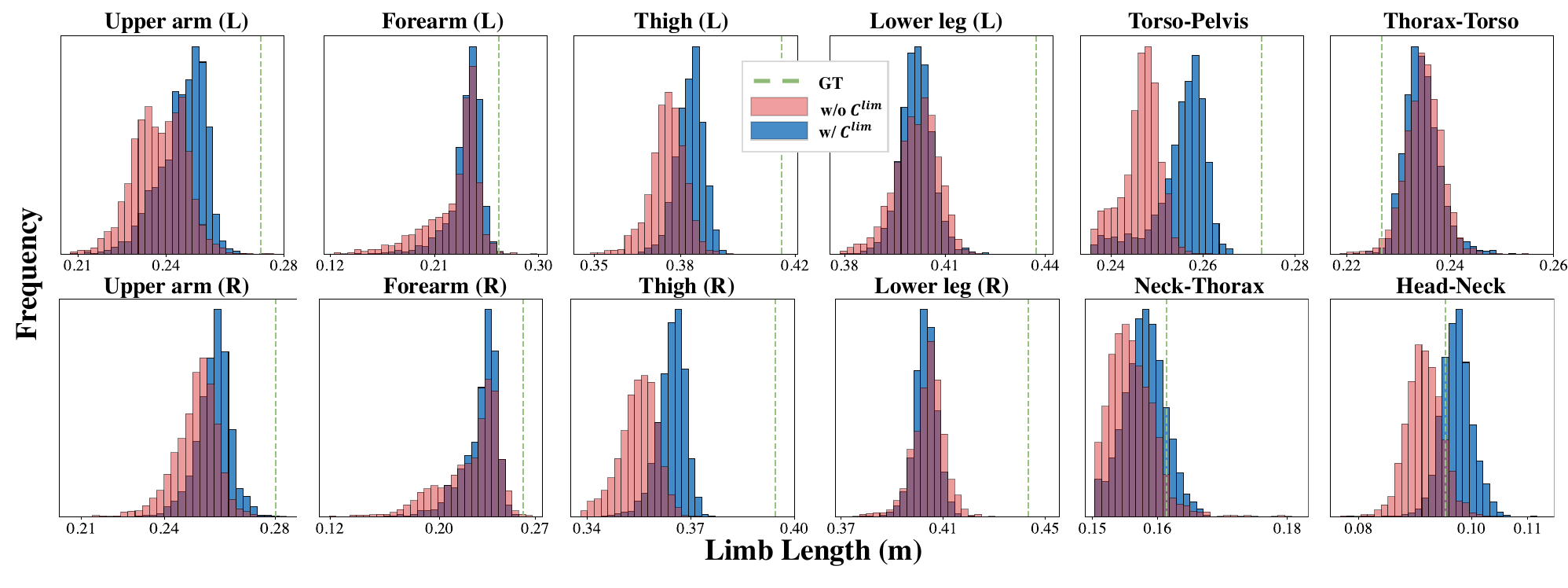}
\caption{Limb-length distribution for a single subject by histogram. The error within 5 cm can be treated as correct. With $C^{lim}$, more accurate limb-length and less variance are observed, as the distribution moves towards the GT and is more concentrated.}
\label{Limb}
\end{figure}

\subsection{Analytics}
\label{Analytics}

\subsubsection{Ablation Study.} 
As shown in Table~\ref{tab:ablation}, we perform the ablation study to verify the effectiveness of each component from the following 4 perspectives. (1) The vanilla diffusion design without any radar guidance can reduce the joint error by modeling pose distribution and refine deformed poses. (2) The effectiveness of the global-local radar context modules is verified by the performance gain of 4.8\% (MPJPE) and 5.5\% (PA-MPJPE). Both modules extract clean radar information with a different focus (local radar PCs or global PC features). (3) The effectiveness of the structure-motion consistency modules are verified by the performance gain of 5.3\% (MPJPE) and 6.1\% (PA-MPJPE). (4) The necessity of each module is further proved, as the elimination of different modules leads to a performance drop. Specifically, the performance drop is more significant when removing the limb-length module ($C^{lim}$) and temporal motion consistency module ($C^{tem}$), as the pose instability causes great pose inaccuracy.

\subsubsection{Effect of Global-local Radar Context.} To demonstrate the effectiveness of joint-wise feature extraction using the joint feature template, we compare our design with an alternative in Table~\ref{tab:global-local}: to directly generate PC feature guidance using the transformer without any template. Our designed joint feature guidance significantly outperforms the PC feature guidance, as the PC features are easily affected by the radar's miss-detection. To demonstrate the effectiveness of dynamic anchors of LRC, in Table~\ref{tab:global-local} we compare our design with static anchor design~\cite{xue2021mmmesh} using coarse estimated human poses $\tilde{H}$. Our design has better performance and is more suitable for diffusion-based local PC selection.

\subsubsection{Effects of Temporal Motion Consistency.} In Figure~\ref{exp2} (a), we observe a mixing of incorrect skeletons within the correct skeleton timeline with PointTrans.~\cite{zhao2021point}, e.g. chest-expanding and hand-waving. With mmDiff, smooth motion patterns are ensured based on human behavioral prior knowledge. Additionally, our proposed mmDiff can mitigate the uncertain locations of legs and arms (caused by low resolution), demonstrating enhanced pose accuracy. Furthermore, for throwing and kicking actions, our method is more stable compared to RGB-extracted ground truth, as camera-based HPE suffers from self-occlusion. In Figure~\ref{exp2} (b), we further apply the Average Keypoint Velocity (AKV) to quantitatively measure the pose stability. AKV is defined as $E_{ i=[0..17]}(||J_t^i - J^i_{t-1}||^2)$, which measures the average inter-frame joint moving distance, indicating the motion energy level and pose stability. The proposed mmDiff demonstrates enhanced pose stability by minimizing joint vibration and avoiding abrupt pose changes.

\begin{table}[!t]
\centering
\caption{Model efficiency evaluated on mmBody. We compare mmDiff's diffusion training in phase two with the benchmark method P4Transformer~\cite{fan2021point}. Extra computational resources of our designed modules for one diffusion step are illustrated.}
\renewcommand{\arraystretch}{1.0}
\adjustbox{width=.8\linewidth}{
    \begin{tabular}{p{11em} p{5em} p{8em} p{6em} p{6em} p{6em}}
    \toprule
    \multicolumn{1}{c}{\textbf{Modules}} & \textbf{Input} & \textbf{Input Size} & \textbf{Latency} & $\mathbf{\#}$\textbf{Params.} & \textbf{GFLOPs.} \\
    \midrule[1.5pt]
    P4Transformer~\cite{fan2021point} & $R$ & $(N, 6)$ & 40.48 ms & 128.00M  & 43.50 \\
    \midrule
    Diffusion model (D) & $\hat{H}_{k}$ & $(17, 3)$ & 7.59 ms & 1.03M & 0.03 \\
    Global context (G)& $F^{j}$ & $(17, 64)$ & 0.36 ms  & 0.02M & 0.01 \\ 
    Local context (L)& $R, \hat{H}_{k}$ & $(N, 6), (17, 3)$ & 2.00 ms & 0.19M & 0.40 \\
    Motion consistency (T)& $\{\Tilde{H}_{t-i}\}_{i=1}^{\Delta t}$ & $(\Delta t, 17, 3)$ & 1.74 ms  & 15.98M & 0.19 \\
    Limb consistency (S) & $F^{j}$ & $(17, 64)$ & 0.16 ms  & 1.29M & 0.02 \\
    (D+G+L+T+S) & /  & /    & 11.85 ms & 18.51M & 0.62 \\
    \bottomrule
    \end{tabular}%
    }
\label{tab:efficiency}
\end{table}

\subsubsection{Effects of Structural Limb-length Consistency.} To validate the effectiveness of the limb-length consistency, we compare mmDiff w/ and w/o the module ($C^{lim}$) and plot the histograms that indicate the limb-length distribution in Figure \ref{Limb}. The ground truth limb length remains constant for all limbs. We observe both reduced limb-length error (in the estimated arms, legs, and spline) and variance (in forearms and lower legs), indicating enhanced pose accuracy and structural stability. We argue that more accurate limb-length can lead to more accurate pose estimation. Qualitatively in figure \ref{exp2}, the variant height and arm’s length for chest expansion and hand raising are mitigated with mmDiff. However, as in Table~\ref{tab:maintable}, SLC has limitations when the sensing distance is long (e.g., in Lab 2) due to harder limb-length estimation.

\subsubsection{Model Efficiency.} We provide the efficiency analysis in Table~\ref{tab:efficiency}, where our proposed modules require substantially little computational resources during phase-two diffusion training and inference. All designed modules have a latency of less than 2ms, demonstrating outstanding efficiency in supporting the iterative diffusion process. Meanwhile, the model's parameters of our designed modules are relatively small compared to the P4Transformer~\cite{fan2021point}, and require minimum computation complexity, as reflected by GFLOPs. It demonstrates the potential for applications like robotics, the Internet of Things, and edge computing.

\section{Conclusion}\label{sec:conclusion}
This paper proposes mmDiff as a human pose estimation (HPE) method for RF-vision, a conditional diffusion model designed to generate accurate and stable human poses from noisy mmWave radar point clouds. The proposed modules demonstrate enhanced robustness in handling radar's miss-detection and signal inconsistency when extracting the guiding features. Compared to the state-of-the-art methods, the proposed mmDiff demonstrates better accuracy and stability in mmWave HPE, showing the viability of radar-based HPE to deal with low illumination or haze. However, it also has limitations worth future studies, including slightly inferior performance compared to RGB-D and increased noise in the presence of multiple sensing targets.

\section*{Acknowledgements}
This research is supported by the National Research Foundation of Singapore under its Medium-Sized Center for Adavnced Robotics Technology Innovation, and Ministry of Education of Singapore under ACRF Tier 1 Grant RG 64/23.

{
    \small
    \bibliographystyle{splncs04}
    
}


\end{document}